\documentclass[conference]{IEEEtran}
\usepackage{cite}
\usepackage{graphicx} % Required for inserting images
\usepackage{tabularx}
\usepackage{booktabs}
\usepackage{fancyvrb}
\usepackage{tikz}
\usetikzlibrary{positioning, arrows.meta, calc}
\usepackage{xcolor}
\usepackage[most]{tcolorbox}
\usepackage{amsmath}
\usepackage{pdflscape}
\usepackage{amssymb}
\usepackage{caption}
\definecolor{custompurple}{HTML}{9F7FAE}
\usepackage{tikz}
\usepackage{booktabs}
\usepackage{multirow}
\usepackage{url}
\usepackage{fvextra}
\usepackage{float}

\usepackage{caption}
\usetikzlibrary{
    positioning,
    arrows.meta
}

\tikzset{
    process/.style={
        rectangle,
        rounded corners=2pt,
        draw,
        minimum height=0.8cm,
        text width=2.2cm,
        align=center,
        font=\scriptsize
    },
    episodic/.style={
        process,
        draw=orange!80!black,
        fill=orange!8
    },
    model/.style={
        process,
        font=\scriptsize\bfseries
    },
    arrow/.style={
        -{Latex[length=2mm]},
        thin
    }
}

\title{Can LLMs Reason Over Long Horizons? An Empirical Evaluation of Context Strategies for Longitudinal Clinical Reasoning}

\author{\IEEEauthorblockN{Taye Akinrele\IEEEauthorrefmark{1},
Noorbakhsh Amiri Golilarz\IEEEauthorrefmark{2},
Subash Neupane\IEEEauthorrefmark{3},
Sudip Mittal\IEEEauthorrefmark{4},
Shahram Rahimi\IEEEauthorrefmark{5}
}
 
\IEEEauthorblockA{Department of Computer Science,
The University of Alabama\\
\{toakinrele\IEEEauthorrefmark{1},
namirigolilarz\IEEEauthorrefmark{2},
sneupane4\IEEEauthorrefmark{3},
smittal1\IEEEauthorrefmark{4},
srahimi1\IEEEauthorrefmark{5}\}@ua.edu}
}

\begin{document}

\maketitle

\begin{abstract}
Longitudinal clinical reasoning requires large language models (LLMs) to identify and integrate relevant evidence distributed across extended patient histories. Although long-context models can process increasingly large amounts of information, providing more history does not necessarily make relevant evidence more accessible or improve reasoning. We compare five context strategies (Full, Recent, Episodic, Semantic, and Hybrid) on \textit{MedLoCoMo} across four open-weight LLMs, examining answer correctness, robustness to query-evidence distance, and abstention on questions with unsupported premises. \textit{Episodic} and \textit{Hybrid} generally achieve the strongest overall accuracy, while \textit{Recent Context} degrades most as supporting evidence becomes more distant; \textit{Episodic} and \textit{Hybrid} maintain the highest accuracy at long distances. Analysis of adversarial questions further shows that strong performance on answerable questions does not necessarily translate to successful abstention when the available history does not support the requested conclusion. These findings show that reliable longitudinal reasoning depends not only on how much history an LLM can access, but critically on how relevant evidence is selected and presented for reasoning.
\end{abstract}

\section{Introduction} \label{sec:introduction}

Clinical decisions often depend on evidence accumulated across a patient's history rather than information contained within a single encounter \cite{zhang2026medlocomo}. Diagnostic findings, treatments, laboratory results, medication changes, and clinical decisions may be distributed across multiple hospitalizations and outpatient visits spanning months or years \cite{hu2026medilongchat}. Longitudinal clinical reasoning therefore requires systems to identify and connect relevant evidence across time to understand disease progression, recurring symptoms, treatment changes, and prior care decisions.

With the emergence of large language models (LLMs) with extended context windows, a straightforward approach to longitudinal clinical reasoning is to provide the model with a patient's full available history. In principle, this keeps relevant evidence from prior encounters accessible during reasoning. However, the ability to accommodate long contexts does not guarantee that all information within them will be used effectively. MedOdyssey demonstrates that long-context LLMs remain sensitive to the position of relevant medical information, exhibiting a lost-in-the-middle effect as context grows \cite{fan2025medodyssey}. LongMemEval similarly reports substantial degradation over long multi-session histories, particularly for multi-session reasoning, temporal reasoning, and knowledge updating \cite{wu2025longmemeval}. In longitudinal clinical settings, cross-admission reasoning remains particularly challenging \cite{zhang2026medlocomo}. These findings suggest that increasing context capacity alone does not solve the problem of identifying and reasoning over relevant evidence distributed across long histories.

An alternative is to selectively construct the context presented to the model. Memory-augmented systems can retrieve information at different levels of abstraction, such as individual experiences or consolidated facts \cite{sumers2024coala}. However, existing memory systems often combine representation, retrieval, consolidation, updating, and context construction within a single architecture, making it difficult to isolate how the context provided to the model affects reasoning performance. This motivates a more fundamental question: \emph{how should longitudinal clinical information be selected and organized for effective long-horizon reasoning?}

\begin{figure}[!t]
    \centering
    \includegraphics[width=0.98\linewidth]{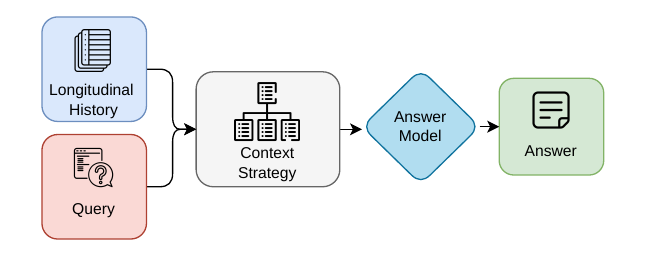}
    \caption{Overview of the experimental framework. Given a longitudinal patient history and query, a context strategy determines the information provided to the answer model. We compare Full Context, Recent Context, Episodic, Semantic, and Hybrid strategies under the same underlying histories and queries.}
    \label{fig:memory_intro}
\end{figure}

This study investigates this question through a controlled empirical evaluation on MedLoCoMo \cite{zhang2026medlocomo}. We compare five \emph{context strategies}: Full Context, Recent Context, Episodic, Semantic, and Hybrid. These strategies range from direct access to longitudinal history to relevance-based retrieval of encounter-level and fact-level information. We examine how context strategy affects reasoning across clinical tasks, robustness to increasing query-evidence distance, and abstention when the available evidence does not support the requested conclusion (illustrated in Fig.~\ref{fig:memory_intro}). We address three research questions:

\begin{itemize}
    \item \textbf{RQ1:} How do different context strategies affect long-horizon reasoning performance, and how does their effectiveness vary across reasoning tasks?
    \item \textbf{RQ2:} How does long-horizon reasoning performance change as the distance between relevant evidence and the query increases, and how does this degradation differ across context strategies?
    \item \textbf{RQ3:} How does context strategy affect abstention when evidence is insufficient?
\end{itemize}

% \noindent Our primary contributions are:

\begin{tcolorbox}[
    enhanced,
    colback=gray!3,
    colframe=gray!45,
    boxrule=0.6pt,
    arc=1.5mm,
    left=5pt,
    right=5pt,
    top=7pt,
    bottom=5pt,
    title=\textbf{Contributions},
    coltitle=white,
    fonttitle=\bfseries,
    attach boxed title to top left={
        xshift=5mm,
        yshift*=-\tcboxedtitleheight/2
    },
    boxed title style={
        colback=gray!70,
        colframe=gray!70,
        boxrule=0pt,
        arc=1mm,
        left=4pt,
        right=4pt,
        top=2pt,
        bottom=2pt
    }
]
\begin{itemize}
    \item We provide a controlled evaluation of five context strategies for longitudinal clinical reasoning, enabling direct comparison of long-context, recency-based, and retrieval-based approaches. 

    \item We show that access to more longitudinal history does not necessarily improve reasoning performance: retrieval-based strategies, particularly Episodic and Hybrid, generally provide stronger performance despite operating over substantially less context.

    \item We demonstrate that context strategy affects multiple dimensions of long-horizon reasoning, including answer correctness, performance on distant evidence, and abstention when the available history does not support the requested conclusion.
\end{itemize}
\end{tcolorbox}

The remainder of this paper is organized as follows. Section~\ref{sec:related_work} reviews related work, followed by the problem statement in  Section~\ref{sec:problem_statement}. Section~\ref{sec:methodology} describes the methodology and experimental setup. Section~\ref{sec:result} presents and discusses the results, and Section~\ref{sec:conclusion} concludes the paper.

\section{Related Work}\label{sec:related_work}
Longitudinal clinical reasoning requires models to integrate information distributed across repeated patient encounters, including changes in diagnoses, medications, laboratory results, and treatments. Recent medical benchmarks show that this remains difficult for LLMs. MedLoCoMo \cite{zhang2026medlocomo} evaluates single- and cross-admission reasoning over longitudinal patient histories grounded in MIMIC-IV, while LongMedBench \cite{chen2026longmedbench} evaluates factual retrieval, temporal reasoning, and clinical decision-making over longitudinal EHRs. Related work such as MediLongChat \cite{hu2026medilongchat} similarly evaluates reasoning across extended medical dialogues. These studies demonstrate that access to a patient's history alone does not guarantee effective longitudinal reasoning, particularly when relevant evidence is distributed across encounters.

This challenge is closely related to research on long-term LLM memory and long-context reasoning. LongMemEval demonstrates substantial performance degradation in information extraction, multi-session reasoning, temporal reasoning, knowledge updating, and abstention over extended histories \cite{wu2025longmemeval}. Similarly, the “lost-in-the-middle” phenomenon shows that information can become difficult for LLMs to use when it occurs between the beginning and end of a long context, even when the information remains within the model's context window \cite{liu2024lostinthemiddle}. These findings motivate approaches that selectively construct the context provided to an LLM rather than relying exclusively on full-context prompting.

Memory-augmented approaches provide one mechanism for constructing such contexts. CoALA distinguishes episodic memory for past experiences, semantic memory for consolidated knowledge, procedural memory for learned procedures, and working memory for information actively used during reasoning \cite{sumers2024coala}. Systems such as MemoryBank \cite{zhong2024memorybank}, MemGPT \cite{packer2023memgpt}, and AdMem \cite{wang2026admem} illustrate different approaches to maintaining and retrieving information beyond an LLM's immediate context. Generative agents, for example, retrieve past experiences according to relevance, recency, and importance and construct higher-level reflections \cite{park2023generative}. More recent work compares semantic, episodic, and procedural memory strategies for long-context question answering, providing evidence that memory design can affect reasoning performance \cite{terranova2025memory}. However, the evaluated memory functions are not independently controlled: episodic and procedural components are layered over semantic retrieval, while working memory remains fixed. MemTrace \cite{deng2026memtrace} further emphasizes the importance of distinguishing failures in accessing relevant information from failures in reasoning after that information has been retrieved.

Taken together, prior work suggests that long-horizon performance depends not only on the amount of history available to an LLM, but also on how that history is selected, organized, and provided during reasoning. We refer to these alternatives as \emph{context strategies}. Existing clinical studies largely evaluate long-context prompting, retrieval methods, or complete memory architectures, making it difficult to compare direct-context and memory-based approaches under a common experimental setting. In particular, controlled evidence remains limited on whether different context strategies are better suited to different forms of longitudinal clinical reasoning, how their effectiveness changes as supporting evidence becomes more distant, and how they affect model behavior when the available history does not support the requested conclusion.

% This study addresses this gap through a controlled comparison of five context strategies: full context, recent context, episodic memory, semantic memory, and hybrid episodic-semantic memory. Full and recent context provide direct-context baselines, while episodic, semantic, and hybrid memory provide retrieval-based strategies that differ in the form of historical information made available to the reasoning model. By evaluating these strategies under a common framework, we examine how context strategy affects performance across clinical reasoning tasks, how performance changes with evidence distance, and whether generated answers remain grounded in the evidence provided to the model.

\section{Problem Statement}\label{sec:problem_statement}

Long-horizon reasoning requires a system to answer questions using information accumulated over an extended history. Let  $\mathcal{H}=\{S_1,S_2,\ldots,S_T\}$ denote a longitudinal history, where each $S_t$ represents information observed at a particular point in time, and let $q$ denote a query about that history. The goal is to produce an answer $\hat{y}$ supported by information contained in $\mathcal{H}$.

As the history grows, relevant evidence may constitute only a small portion  of the available information and may be separated from the query by many intervening observations. We therefore define a \emph{context strategy} $C$ as a function that determines how information from the longitudinal history is selected, organized, and provided to the answer model. Given a history $\mathcal{H}$ and query $q$,

\begin{equation}
    E_q = C(q,\mathcal{H}), \qquad
    \hat{y}=g(q,E_q),
\end{equation}

where $E_q$ is the context provided to the answer model $g$.

Different context strategies may preserve, prioritize, or retrieve different parts of the longitudinal history. Consequently, providing more history does not necessarily provide more useful evidence for reasoning. This work investigates how context strategy affects answer correctness, robustness to evidence distance, and abstention under insufficient evidence.

\section{Methodology} \label{sec:methodology}

This study evaluates how different context strategies affect an LLM's ability to answer questions requiring information distributed across long, multi-session histories. We compare five strategies: \textit{full\_context}, \textit{recent\_context}, \textit{episodic}, \textit{semantic}, and \textit{hybrid} (\textit{episodic + semantic}). MedLoCoMo \cite{zhang2026medlocomo} serves as the primary benchmark for longitudinal clinical reasoning. To assess whether the observed effects generalize beyond the clinical domain, we additionally evaluate the same strategies on LongMemEval \cite{wu2025longmemeval}, a general-domain benchmark for long-term conversational memory (results are reported in Section~\ref{sec:longmemeval}).

\subsection{Datasets}

\textbf{MedLoCoMo:} We evaluate on MedLoCoMo, a longitudinal clinical question-answering benchmark comprising 100 patient timelines organized as chronologically ordered hospital admissions. Each patient record contains admission-level conversations and evidence-linked question-answer pairs spanning single-admission and cross-admission reasoning. We converted the native per-patient representation into a common flat schema containing the question, answer, longitudinal history, temporal metadata, and identifiers of the supporting admissions, allowing all context strategies to operate over the same underlying patient histories.

For the experiments, we sampled 1,000 questions and used the same question set across all context strategies and answer models. The sample was
approximately balanced across the seven combinations of MedLoCoMo's \texttt{scope} and \texttt{question\_type} fields, as shown in Table~\ref{tab:question_dist}. The five answerable categories comprise medical reasoning, care-plan rationale, longitudinal progression, cross-admission comparison, and frequency pattern, while the two adversarial categories contain questions with unsupported premises. Together, the adversarial categories account for 286 of the 1,000 evaluation questions.

For answerable questions, \texttt{answer\_session\_ids} records the admissions containing the supporting evidence. For adversarial questions, we set \texttt{answer\_session\_ids} to the empty set because the evidence field supplied by MedLoCoMo identifies the near-miss admissions used to construct the unsupported question rather than genuine supporting evidence.

The patient histories are substantially longer than the evidence budget available to the retrieval-based context strategies. As summarized in Table~\ref{tab:dataset_stats}, a patient timeline contains an average of 29.5 admission sessions and 1,662.3 dialogue turns, corresponding to
approximately 35,073 tokens under the token-length estimator used in our experimental pipeline. The longest timeline contains 64 sessions and 3,522 turns, with an estimated length of 74,255 tokens. These statistics are computed over the 100 patient timelines, with each patient counted once rather than once per question.

\begin{table}[t]
\centering
\caption{Distribution of question types in the 1,000-question evaluation
subset.}
\label{tab:question_dist}
\small
\begin{tabular}{llrr}
\toprule
\textbf{Question Type} & \textbf{Scope} &
\textbf{Count} & \textbf{\%} \\
\midrule
Medical reasoning          & Single & 142 & 14.2 \\
Care-plan rationale        & Single & 143 & 14.3 \\
Adversarial                & Single & 143 & 14.3 \\
Longitudinal progression   & Cross  & 143 & 14.3 \\
Cross-admission comparison & Cross  & 143 & 14.3 \\
Frequency pattern          & Cross  & 143 & 14.3 \\
Adversarial                & Cross  & 143 & 14.3 \\
\midrule
\textbf{Total} & & \textbf{1,000} & \textbf{100.0} \\
\bottomrule
\end{tabular}
\end{table}

\begin{table}[t]
\centering
\caption{Longitudinal context statistics for the patient timelines.
Token counts are estimated using the same approximation employed for
experimental evidence budgeting.}
\label{tab:dataset_stats}
\small
\begin{tabular}{lrrr}
\toprule
\textbf{Statistic} &
\textbf{Sessions} &
\textbf{Turns} &
\textbf{Est. Tokens} \\
\midrule
Mean   & 29.5 & 1,662.3 & 35,072.6 \\
Median & 28.0 & 1,623.0 & 34,099.0 \\
Min    & 17   & 913     & 20,451 \\
Max    & 64   & 3,522   & 74,255 \\
\bottomrule
\end{tabular}
\end{table}

\begin{figure*}[h]
    \centering
    \includegraphics[width=\linewidth]{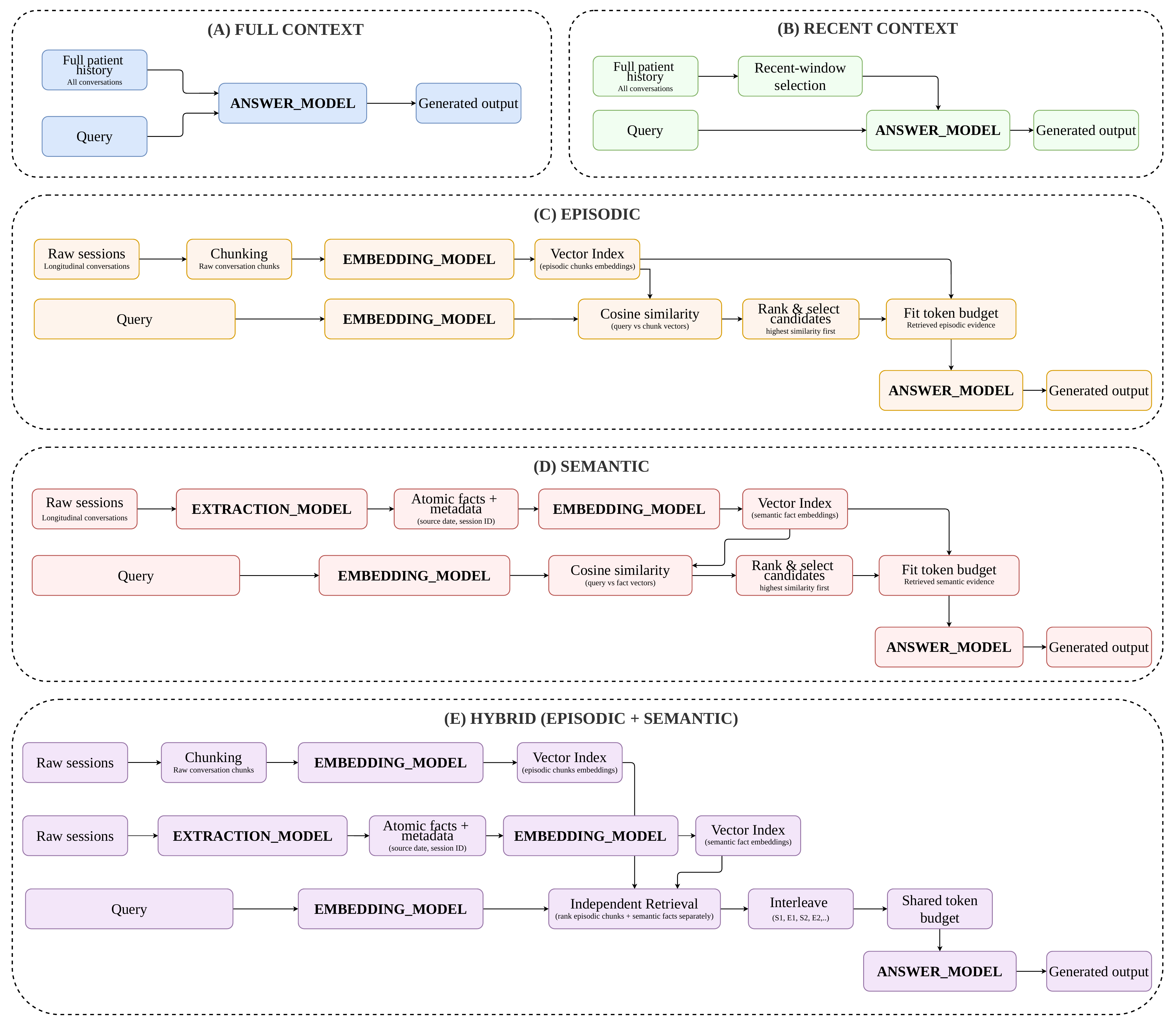}
    \caption{Context strategies evaluated for longitudinal clinical reasoning. Full and Recent provide direct-context baselines, while Episodic, Semantic, and Hybrid use relevance-based retrieval over encounter-level evidence, patient-specific facts, or their combination, respectively.}
    \label{fig:memory_pipelines}
\end{figure*}

\subsection{Context Strategies}

We evaluate five context strategies (Fig.~\ref{fig:memory_pipelines}): Full, Recent, Episodic, Semantic, and Hybrid. Full uses the available longitudinal history up to the model's context limit, while the remaining strategies operate under a common evidence budget.

\begin{itemize}

    \item \textbf{\textit{Full context:}} The longitudinal history is provided verbatim without relevance-based retrieval or compression. All sessions are included when they fit within the model's context window; otherwise, the history is truncated to the available context limit.

    \item \textbf{\textit{Recent context:}} The most recent raw sessions are retained until the evidence budget is exhausted, without relevance-based retrieval or compression. This provides a budget-matched recency baseline.

    \item \textbf{\textit{Episodic:}} The original history is divided into overlapping encounter-level text chunks. Each chunk is embedded independently and ranked by cosine similarity to the query, with the highest-ranked chunks selected within the evidence budget.

    \item \textbf{\textit{Semantic:}} Each session is transformed through an LLM-based extraction step into self-contained patient-specific facts with source-session and temporal metadata. Facts are embedded independently, ranked by cosine similarity to the query, and selected within the evidence budget.

    \item \textbf{\textit{Hybrid:}} Episodic chunks and semantic facts are maintained in separate memory stores and retrieved independently. Similarity-ranked candidates from both stores are interleaved by rank and selected under a shared evidence budget.

\end{itemize}

\subsection{Fact Extraction and Caching}

% We extracted semantic facts independently from each session using \textbf{Qwen3} \cite{yang2025qwen3}. Each session was processed in a single LLM call with a structured JSON output format (\texttt{{"facts": [...]}}). A worked example was included in the extraction prompt to encourage the generation of standalone factual statements rather than reproducing the original conversational structure. Extracted facts were cached to disk by session identifier and extraction model, allowing the same semantic representation to be reused across questions and experimental conditions. This avoids repeated LLM extraction calls for sessions that recur across the dataset, reducing preprocessing cost while keeping the semantic representations consistent across conditions.

Semantic memory was constructed by extracting patient-specific factual statements from each session independently using Qwen3-4B~\cite{yang2025qwen3}. We used LLM-based extraction because clinical conversations often express temporally and causally related facts that entity-level extraction does not capture. Each session was processed in a single structured-output call that returned at most ten self-contained facts, preserving medications, diagnoses, procedures, findings, temporal information, negation, and uncertainty. Extracted facts were cached by session and extraction model, so all questions and conditions referencing the same session used an identical semantic representation.

\begin{figure}[!t]
\centering
\scriptsize

% ============================================================
% QUESTION
% ============================================================

\begin{tcolorbox}[
    width=\linewidth,
    colback=gray!5,
    colframe=black!40,
    boxrule=0.4pt,
    arc=1mm,
    left=3pt,right=3pt,top=2pt,bottom=2pt
]
\textbf{Question:} During the hospitalization from 2155-12-13 to
2155-12-16, why was Coumadin held during the initial post-procedure period?

\textbf{Gold:} Risk of bleeding from the tube site.
\end{tcolorbox}

\vspace{1mm}

% ============================================================
% ROW 1: FULL + RECENT
% ============================================================

\begin{minipage}[t]{0.49\linewidth}
\begin{tcolorbox}[
    colback=blue!3,
    colframe=blue!65!black,
    colbacktitle=blue!65!black,
    coltitle=white,
    title=\textbf{Full Context},
    fonttitle=\bfseries,
    boxrule=0.5pt,
    arc=1mm,
    left=3pt,right=3pt,top=1pt,bottom=1pt,
    height=2.8cm,
    valign=top
]
\textbf{Raw longitudinal history}

\smallskip
\textit{2155-12-13:}
``We're holding your Coumadin for now.
The tube site could bleed...''

\smallskip
\textit{[Earlier and later history]}

\vfill
\textbf{Output:} Coumadin was held to prevent
bleeding at the nephrostomy tube site.
\end{tcolorbox}
\end{minipage}
\hfill
\begin{minipage}[t]{0.49\linewidth}
\begin{tcolorbox}[
    colback=green!3,
    colframe=green!55!black,
    colbacktitle=green!55!black,
    coltitle=white,
    title=\textbf{Recent Context},
    fonttitle=\bfseries,
    boxrule=0.5pt,
    arc=1mm,
    left=3pt,right=3pt,top=1pt,bottom=1pt,
    height=2.8cm,
    valign=top
]
\textbf{Most recent raw history}

\smallskip
\textit{2155-12-19:}
``Coumadin was restarted...''

\textit{2155-12-13:}
``The tube site could bleed...''

\vfill
\textbf{Output:} Coumadin was held to prevent
bleeding at the nephrostomy tube site.
\end{tcolorbox}
\end{minipage}

\vspace{1mm}

% ============================================================
% ROW 2: EPISODIC + SEMANTIC
% ============================================================

\begin{minipage}[t]{0.49\linewidth}
\begin{tcolorbox}[
    colback=orange!3,
    colframe=orange!75!black,
    colbacktitle=orange!75!black,
    coltitle=white,
    title=\textbf{Episodic},
    fonttitle=\bfseries,
    boxrule=0.5pt,
    arc=1mm,
    left=3pt,right=3pt,top=1pt,bottom=1pt,
    height=2.8cm,
    valign=top
]
\textbf{Retrieved encounter}

\smallskip
\textbf{[e6]} 2155-12-13

Patient: ``I'm worried about being on
Coumadin.''

Doctor: ``The tube site could bleed.''

\vfill
\textbf{Output:} Coumadin was held to prevent
bleeding at the nephrostomy tube site.
\end{tcolorbox}
\end{minipage}
\hfill
\begin{minipage}[t]{0.49\linewidth}
\begin{tcolorbox}[
    colback=purple!2,
    colframe=purple!50!black,
    colbacktitle=purple!50!black,
    coltitle=white,
    title=\textbf{Semantic},
    fonttitle=\bfseries,
    boxrule=0.5pt,
    arc=1mm,
    left=3pt,right=3pt,top=1pt,bottom=1pt,
    height=2.8cm,
    valign=top
]
\textbf{Retrieved facts}

\smallskip
\textbf{[s111]} Coumadin is temporarily held
because of bleeding risk from the tube.

\textit{Source: 2155-12-13}

\vfill
\textbf{Output:} Coumadin was held because of
bleeding risk from the tube.
\end{tcolorbox}
\end{minipage}

\vspace{1mm}

% ============================================================
% ROW 3: HYBRID
% ============================================================

\begin{tcolorbox}[
    width=\linewidth,
    colback=purple!4,
    colframe=purple!70!black,
    colbacktitle=purple!70!black,
    coltitle=white,
    title=\textbf{Hybrid (Episodic + Semantic)},
    fonttitle=\bfseries,
    boxrule=0.5pt,
    arc=1mm,
    left=3pt,right=3pt,top=1pt,bottom=1pt
]

\begin{minipage}[t]{0.48\linewidth}
\textbf{Retrieved fact}

\textbf{[s111]} Coumadin is being held because
of bleeding risk from the tube.
\end{minipage}
\hfill
\begin{minipage}[t]{0.48\linewidth}
\textbf{Retrieved encounter}

\textbf{[e6]} Doctor: ``We're holding your
Coumadin for now. The tube site could bleed.''
\end{minipage}

\smallskip
\textbf{Output:} Coumadin was held due to the risk of bleeding from the
nephrostomy tube.

\end{tcolorbox}

\caption{Illustrative example of the five context strategies. }
% The clinical question and answer model remain fixed across conditions; only the context constructed from the longitudinal history changes. Full and Recent provide raw history, Episodic retrieves encounter-level evidence, Semantic retrieves extracted patient-specific facts, and Hybrid combines both representations.
\label{fig:context_strategy_example}

\end{figure}

\subsection{Answer Generation}

For each strategy, the retrieved context and query were provided to the answer model in a single structured-output call, as illustrated in Fig. \ref{fig:context_strategy_example}. The model returned a JSON object containing the generated answer and the identifiers of the supporting excerpts. This format was adopted after preliminary testing showed that separately generating and parsing citations was less reliable. Before evaluation, cited identifiers not present in the context provided to the model were discarded, preventing unsupported or hallucinated citations from being credited. Experiments were repeated across multiple answer models while keeping the memory construction, retrieval procedure, prompts, and evaluation protocol fixed.

\subsection{Evaluation}

Evaluation is organized around the three research questions. All conditions are evaluated on the same sampled questions to enable paired comparison.

\paragraph{\textbf{RQ1 - Answer Correctness}}
The primary outcome is end-to-end answer accuracy. A judge LLM from a different model family than the answer model evaluates each generated response against the gold answer as correct or incorrect based on semantic equivalence rather than exact string match, following the LLM-as-judge protocol of Zheng et al.~\cite{zheng2023judging}. For context strategy $m$, accuracy is defined as

{\footnotesize
\begin{equation}
\mathrm{Accuracy}_{m}
=
\frac{1}{N_m}
\sum_{i=1}^{N_m} y_{im},
\end{equation}
where $N_m$ is the total number of benchmark questions evaluated under context strategy $m$, and
\begin{equation}
y_{im} =
\begin{cases}
1, & \text{if the generated answer is judged correct},\\
0, & \text{if the answer is judged incorrect or generation fails}.
\end{cases}
\end{equation}
}

Accuracy is reported both overall and by question type to examine how the effect of context strategy varies across reasoning tasks. Generation failures therefore remain in the denominator and are counted as incorrect, making this an end-to-end measure of system performance. We use Gemma4-26B \cite{gemma2026gemma4} as the primary correctness judge and MedGemma-27B \cite{sellergren2025medgemma} as a domain-specific secondary judge; neither belongs to the model family of any answer model. Both judges receive the question, gold answer, and generated answer and output a binary \textsc{correct}/\textsc{incorrect} label. We report agreement between the two judges using raw agreement and Cohen's $\kappa$.

\paragraph{\textbf{RQ2 - Robustness to Evidence Distance}}
We measure evidence distance by the location of the gold supporting evidence relative to the end of the longitudinal history associated with each question. Specifically, the session gap $d_i$ captures how far back in the history the supporting evidence occurs. For descriptive analysis, questions are grouped into five predefined distance bins (0--2, 3--5, 6--10, 11--20, and 21+ sessions), and end-to-end answer accuracy is reported for each context strategy within each bin. To test whether the relationship between evidence distance and answer correctness differs across context strategies, we fit the logistic regression model

{\footnotesize
\begin{equation}
\operatorname{logit}\!\left[P(Y_{im}=1)\right] = \beta_0 +\sum_{m=2}^{M}\beta_m C_{im} +\beta_d z_i +\sum_{m=2}^{M}\gamma_m(C_{im}z_i),
\end{equation}
}
where $Y_{im}=1$ indicates a correct answer for question $i$ under context strategy $m$, $C_{im}$ denotes the context-strategy indicators, and $z_i$ is the standardized continuous session gap,

{\footnotesize
\begin{equation}
z_i=\frac{d_i-\bar{d}}{s_d},
\end{equation}
}
where $\bar{d}$ and $s_d$ denote the mean and standard deviation of the raw session gaps, respectively. The interaction coefficients $\gamma_m$ test whether the
effect of evidence distance on answer correctness differs between each strategy and the reference strategy. Because each benchmark question is evaluated under multiple context strategies, we compute cluster-robust standard errors at the question level. We use distance bins only for descriptive analysis and visualization; the regression retains session gap as a continuous variable. As a sensitivity analysis, we additionally report distance-stratified accuracy restricted to successful generations.
        
\paragraph{\textbf{RQ3 - Abstention}} For adversarial questions, which are constructed such that the requested conclusion is unsupported by the patient history, we evaluate whether the generated response appropriately rejects the unsupported premise or otherwise indicates that the question cannot be answered from the available evidence. Responses are evaluated against the benchmark's unanswerable gold response using the same semantic correctness judge as RQ1.

\paragraph{\textbf{Supplementary Metrics}} We compute BERTScore F1~\cite{zhang2020bertscore} between generated and gold answers using \texttt{roberta-large} without baseline rescaling, excluding failed or empty generations; coverage is the percentage of questions with a scorable response. For evidence grounding, each cited excerpt or fact is mapped to its source session, and the cited sessions are compared with the gold sessions on answerable questions. Citation coverage is the percentage of questions with at least one valid citation. Precision, recall, and F1 are computed per question (0 when nothing is cited) and then averaged.

\subsection{Implementation}
All experiments were conducted locally on a workstation with an NVIDIA GeForce RTX 5080 GPU, 64\,GB RAM, and a 24-core Intel Core Ultra 9 285K processor running Ubuntu 24.04.4 LTS. Models were served using Ollama\footnote{\url{https://ollama.com}}. We evaluate \texttt{qwen3-8b} \cite{yang2025qwen3}, \texttt{mistral-7b} \cite{jiang2023mistral}, \texttt{llama3.2-3b} \cite{grattafiori2024llama3}, and \texttt{glm4-9b} \cite{glm2024chatglm} as answer models. Full Context uses each model's available context window, whereas Recent Context and the retrieval-based conditions are constrained to a common 7,000-token evidence budget.

Semantic memory is constructed using \texttt{qwen3-4b} and both episodic and semantic memories are embedded using \texttt{all-MiniLM-L6-v2} \cite{wang2020minilm, reimers2019sentencebert} and ranked using cosine similarity. Answer-generation prompts and decoding settings are held fixed across context strategies for each answer model, while retrieval and context construction vary according to the strategy being evaluated.

\section{Results \& Discussion}\label{sec:result}
We report results across four answer models with different context capacities. Qwen and Mistral use 32k-token context windows, while LLaMA and GLM-4 use 128k-token windows. This difference primarily affects Full Context, for which longer histories may be truncated to the model's context limit. Recent, Episodic, Semantic, and Hybrid instead operate under the same evidence budget across models. Full Context results should therefore be interpreted with these differences in context capacity in mind.

\subsection{Overall Effect of Context Strategy}

\begin{figure*}[h]
    \centering
    \begin{minipage}[t]{0.45\textwidth}
        \centering
        \includegraphics[width=\linewidth]{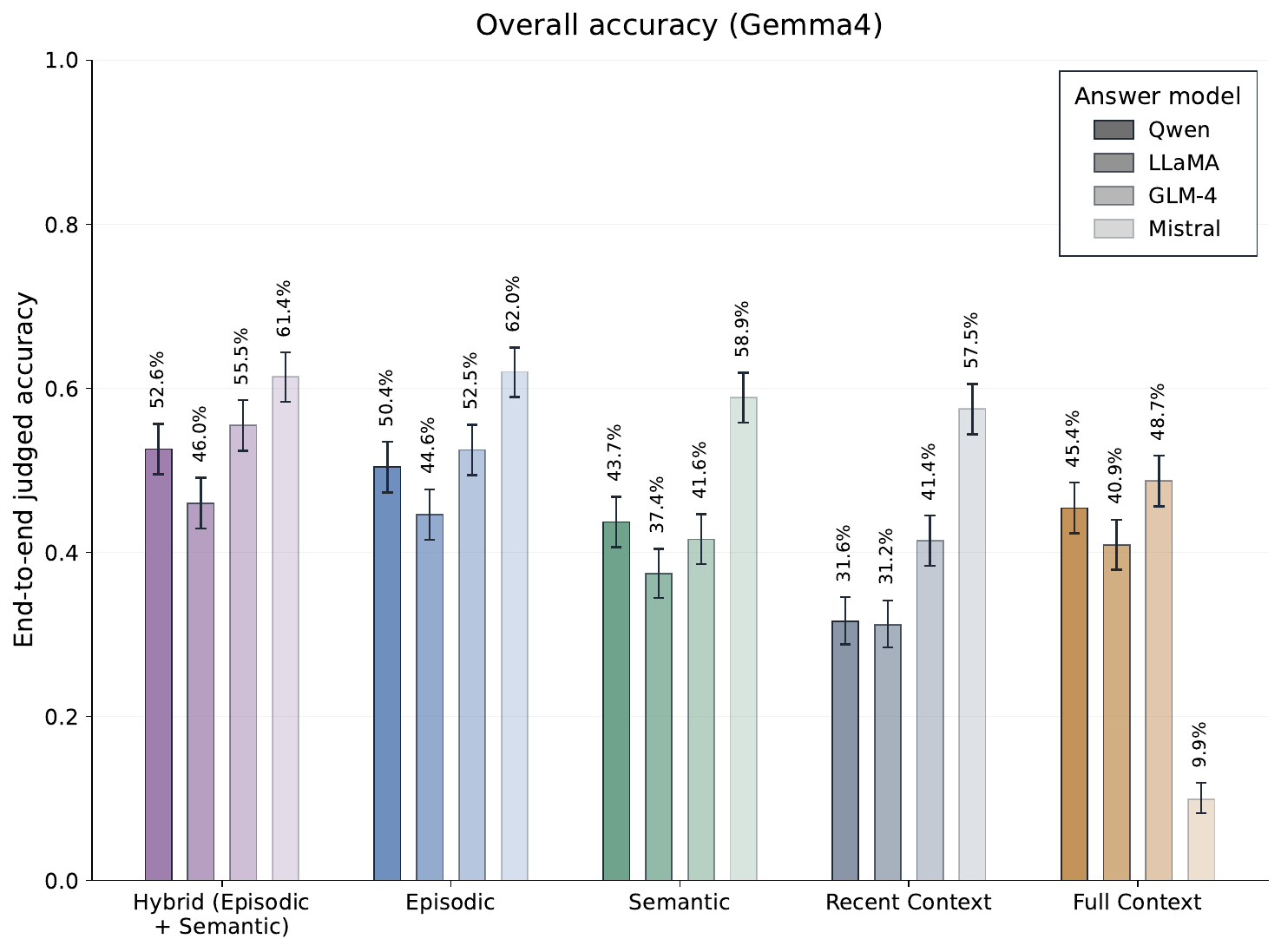}
        \vspace{-2mm}
        \centerline{\small (a)}
    \end{minipage}
    \hfill
    \begin{minipage}[t]{0.45\textwidth}
        \centering
        \includegraphics[width=\linewidth]{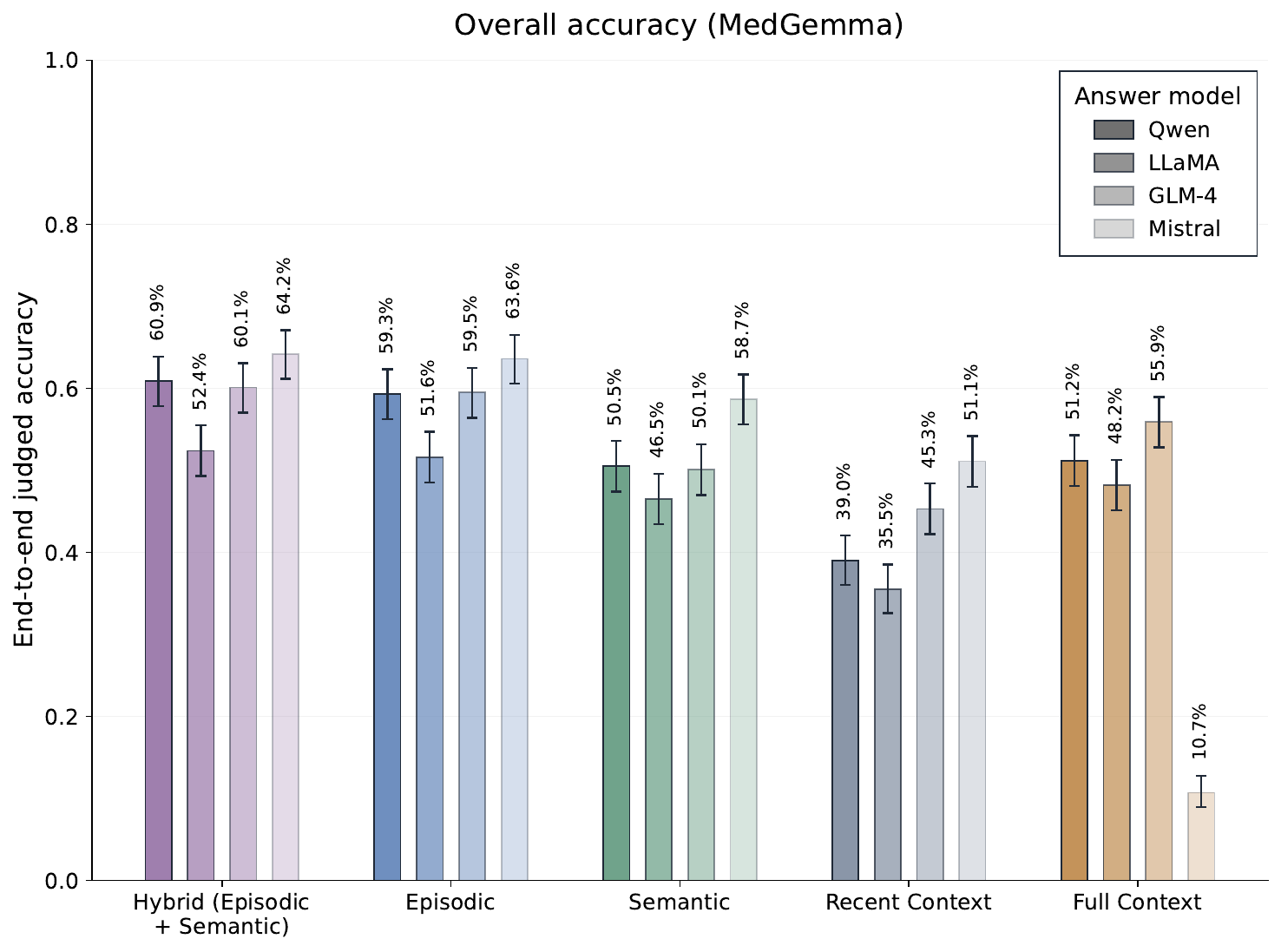}
        \vspace{-2mm}
        \centerline{\small (b)}
    \end{minipage}
    \hfill
    \begin{minipage}[t]{0.45\textwidth}
        \centering
        \includegraphics[width=\linewidth]{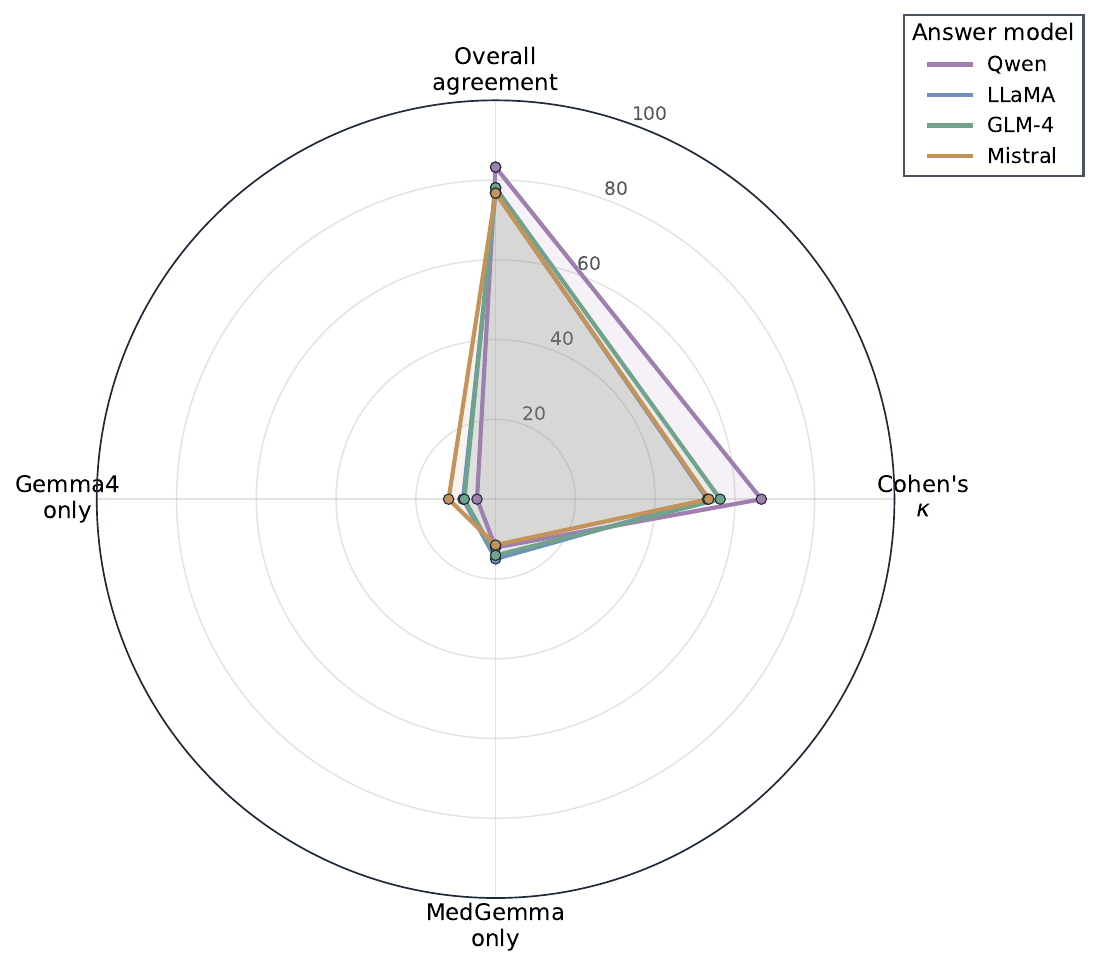}
        \vspace{-2mm}
        \centerline{\small (c)}
    \end{minipage}
    \hfill
    \begin{minipage}[t]{0.45\textwidth}
        \centering
        \includegraphics[width=\linewidth]{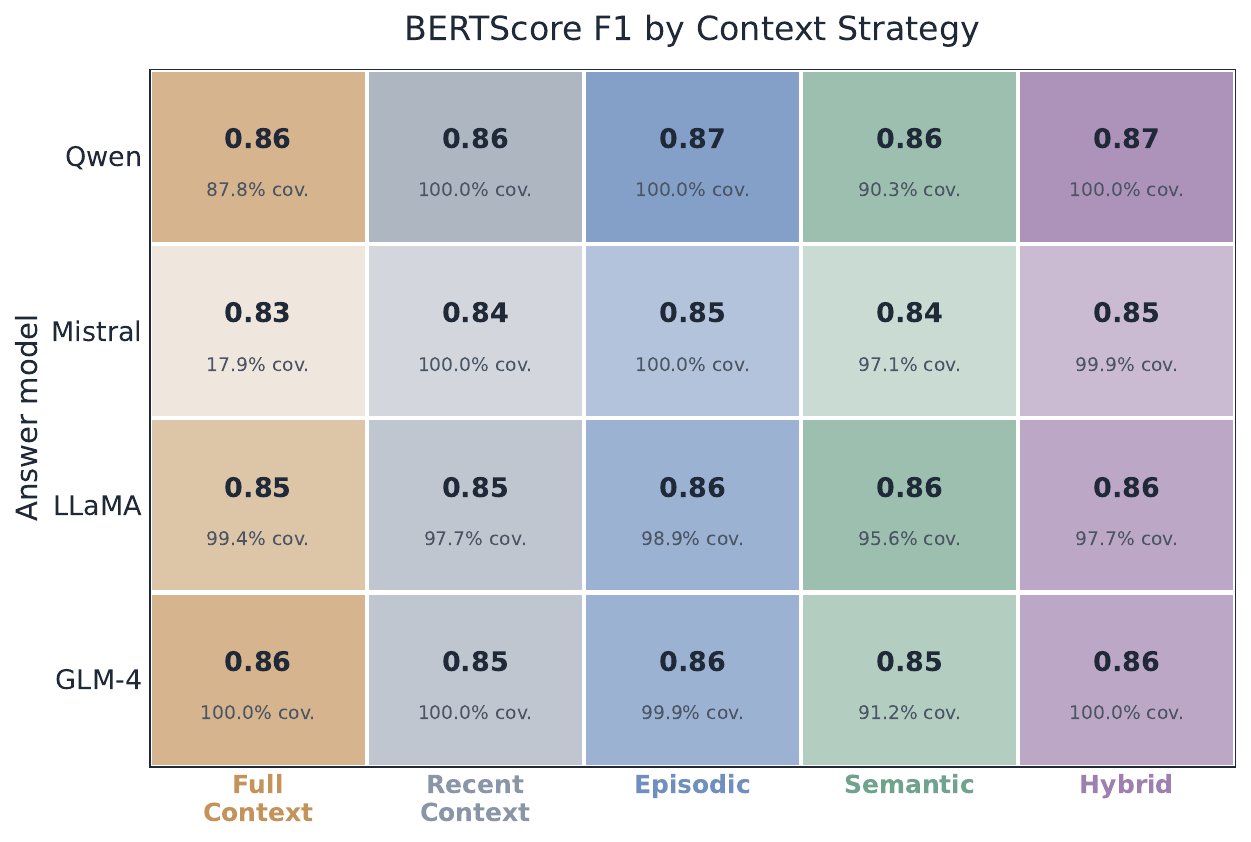}
        \vspace{-2mm}
        \centerline{\small (d)}
    \end{minipage}

    \caption{Overall accuracy and judge sensitivity on MedLoCoMo. (a–b) End-to-end accuracy across context strategies and answer models using Gemma4 and MedGemma as correctness judges, respectively; error bars indicate 95\% Wilson confidence intervals. (c) Agreement between the two correctness judges, including overall agreement, Cohen's $\kappa$, and directional disagreements. (d) BERTScore F1 between generated and reference answers for the primary Gemma4-evaluated runs; percentages within cells indicate the proportion of generations evaluable by BERTScore.}
    \label{fig:medlocomo_overall_accuracy}
\end{figure*}

We first examine how context strategy affects overall answer accuracy on MedLoCoMo. Using \textit{Gemma4} as the primary correctness judge, Hybrid achieves the highest accuracy for Qwen (52.6\%), LLaMA (46.0\%), and GLM-4 (55.5\%), while Episodic performs best for Mistral (62.0\%), closely followed by Hybrid (61.4\%) (Fig.~\ref{fig:medlocomo_overall_accuracy}a). Overall, Hybrid and Episodic provide the strongest aggregate performance, although their relative advantage varies across answer models. We repeat the evaluation using MedGemma as a domain-specific judge  (Fig.~\ref{fig:medlocomo_overall_accuracy}b). The two judges agree on 78.7\% of evaluations, with a Cohen's $\kappa$ of 0.575. MedGemma assigns correct judgments more frequently in disagreements (13.2\%) than Gemma4 (8.1\%), resulting in generally higher absolute accuracy (Fig.~\ref{fig:medlocomo_overall_accuracy}c). Nevertheless, the main strategy-level patterns are preserved, with Hybrid and Episodic remaining among the strongest conditions across models. Mistral Full Context produced a valid answer for only 179 of 1,000 questions due to generation failures; on successful generations its accuracy was 55.3\%, comparable to other strategies.

As a complementary measure of answer similarity, BERTScore F1 varies only modestly across strategies, ranging from 0.83 to 0.87 (Fig.~\ref{fig:medlocomo_overall_accuracy}d). Episodic and Hybrid achieve the highest or near-highest scores across answer models, but the differences are substantially smaller than those observed in judged accuracy. For example, Qwen achieves BERTScore F1 values of 0.86 with Recent Context and 0.87 with Hybrid, despite judged accuracies of 31.6\% and 52.6\%, respectively. This indicates that semantic similarity alone does not capture the context-dependent differences in answer correctness identified by the correctness judge. BERTScore coverage is high for most conditions, with Mistral Full Context as a notable exception (17.9\%).

\begin{table}[t]
\centering
\caption{Session-level citation grounding across context strategies. Coverage denotes the percentage of evaluated responses containing at least one citation that could be mapped to a source session.}
\label{tab:citation_grounding}
\begin{tabular}{lcccc}
\toprule
Strategy & Coverage (\%) & Precision (\%) & Recall (\%) & F1 (\%) \\
\midrule
Full Context   & 44.9 & 22.0 & 27.6 & 21.2 \\
Recent Context & 63.2 & 13.5 & 11.1 & 11.2 \\
Episodic       & 87.5 & \textbf{48.7} & \textbf{50.0} & \textbf{46.1} \\
Semantic       & 79.0 & 35.7 & 40.5 & 34.8 \\
Hybrid         & \textbf{87.9} & 41.3 & 47.9 & 40.4 \\
\bottomrule
\end{tabular}
\end{table}

\begin{table*}[t]
\centering
\caption{\small Accuracy (\%) by context strategy and question type on the MedLoCoMo dataset across four answer models. Each model is evaluated on 1,000 questions: Cross-Admission Adversarial (CA-Adv; $n=143$), Cross-Admission Comparison (CA-Comp; $n=143$), Cross-Admission Frequency Pattern (CA-Freq; $n=143$), Cross-Admission Longitudinal Progression (CA-Long; $n=143$), Single-Admission Adversarial (SA-Adv; $n=143$), Single-Admission Care-Plan Rationale (SA-Care; $n=143$), and Single-Admission Medical Reasoning (SA-Med; $n=142$). Best performance within each answer model and question type is shown in bold.}
\label{tab:rq1_medlocomo}
\setlength{\tabcolsep}{9pt}
\renewcommand{\arraystretch}{1.15}
\resizebox{\textwidth}{!}{
\begin{tabular}{llccccccc}
\toprule
\textbf{Model} & \textbf{Strategy} & \textbf{CA-Adv} & \textbf{CA-Comp} & \textbf{CA-Freq} & \textbf{CA-Long} & \textbf{SA-Adv} & \textbf{SA-Care} & \textbf{SA-Med} \\
\midrule
\multirow{5}{*}{Qwen-3-8B}
& Full Context & \textbf{6.99} & 51.05 & 41.96 & 53.85 & 26.57 & 66.43 & 71.13 \\
& Recent Context & 0.70 & 45.45 & 38.46 & 31.47 & 16.78 & 46.15 & 42.25 \\
& Episodic & 1.40 & 55.24 & \textbf{49.65} & 51.75 & 30.07 & \textbf{88.11} & 76.76 \\
& Semantic & 0.00 & 48.95 & 34.27 & 51.05 & \textbf{37.06} & 72.73 & 61.97 \\
& Hybrid (E+S) & 2.80 & \textbf{56.64} & 47.55 & \textbf{60.14} & 36.36 & 86.71 & \textbf{78.17} \\
\midrule
\multirow{5}{*}{Mistral-7B}
& Full Context & 32.17 & 11.19 & 1.40 & 1.40 & 2.10 & 16.08 & 4.93 \\
& Recent Context & \textbf{59.44} & 53.15 & 53.15 & \textbf{57.34} & 58.74 & 67.13 & 53.52 \\
& Episodic & 46.85 & \textbf{62.24} & \textbf{55.24} & 53.15 & 50.35 & 83.92 & \textbf{82.39} \\
& Semantic & 52.45 & 59.44 & 52.45 & 48.95 & \textbf{72.03} & 67.13 & 59.86 \\
& Hybrid (E+S) & 44.06 & 61.54 & 49.65 & 55.94 & 51.05 & \textbf{90.91} & 76.76 \\
\midrule
\multirow{5}{*}{GLM-4-9B}
& Full Context & 6.29 & 58.74 & 47.55 & \textbf{55.94} & 18.18 & 81.82 & 72.54 \\
& Recent Context & \textbf{42.66} & 46.15 & 34.27 & 34.97 & 24.48 & 64.34 & 42.96 \\
& Episodic & 27.97 & 60.14 & 44.76 & 54.55 & 18.18 & 84.62 & 77.46 \\
& Semantic & 25.17 & 40.56 & 33.57 & 39.86 & 27.27 & 67.13 & 57.75 \\
& Hybrid (E+S) & 30.77 & \textbf{60.84} & \textbf{48.95} & 51.05 & \textbf{30.77} & \textbf{88.11} & \textbf{78.17} \\
\midrule
\multirow{5}{*}{LLaMA-3.2-3B}
& Full Context & 7.69 & \textbf{53.85} & 40.56 & 41.26 & 18.18 & 67.83 & 57.04 \\
& Recent Context & 14.69 & 31.47 & 32.87 & 32.87 & \textbf{27.97} & 44.06 & 34.51 \\
& Episodic & \textbf{30.77} & 48.95 & 38.46 & 38.46 & 23.78 & 66.43 & \textbf{65.49} \\
& Semantic & 27.27 & 37.76 & 36.36 & 38.46 & 26.57 & 47.55 & 47.89 \\
& Hybrid (E+S) & 30.07 & 46.85 & \textbf{45.45} & \textbf{41.96} & 21.68 & \textbf{71.33} & 64.79 \\
\bottomrule
\end{tabular}}
\end{table*}

To examine evidence grounding, we compare the evidence explicitly cited by the model with the annotated supporting sessions. Each cited excerpt or semantic fact is mapped to its source session and compared with the gold evidence-session identifiers. Table~\ref{tab:citation_grounding} reports citation coverage, precision, recall, and F1 for questions with annotated supporting evidence. Episodic achieves the highest citation precision (48.7\%), recall (50.0\%), and F1 (46.1\%), while Hybrid achieves the highest citation coverage (87.9\%) and the second-highest recall (47.9\%). Full and Recent Context show substantially lower citation precision and recall, indicating weaker alignment between cited and annotated evidence sessions.

\subsection{Performance Across Reasoning Tasks}

The effect of context strategy varies across reasoning tasks (Table~\ref{tab:rq1_medlocomo}). No single strategy achieves the highest accuracy across all question types and answer models. For care-plan rationale, Hybrid achieves the highest accuracy for Mistral (90.91\%), GLM-4 (88.11\%), and LLaMA (71.33\%), while Episodic achieves the highest accuracy for Qwen (88.11\%). The highest-performing strategy for cross-admission comparison, frequency-pattern, and longitudinal-progression questions also varies across answer models. Performance also varies substantially on the two adversarial question types. We examine this behavior separately in Section~\ref{sec:abstention} as part of RQ3. Overall, the task-level results demonstrate that the relative effectiveness of context strategies varies across both reasoning tasks and answer models. 

\begin{figure*}[h]
    \centering
    \includegraphics[width=\linewidth]{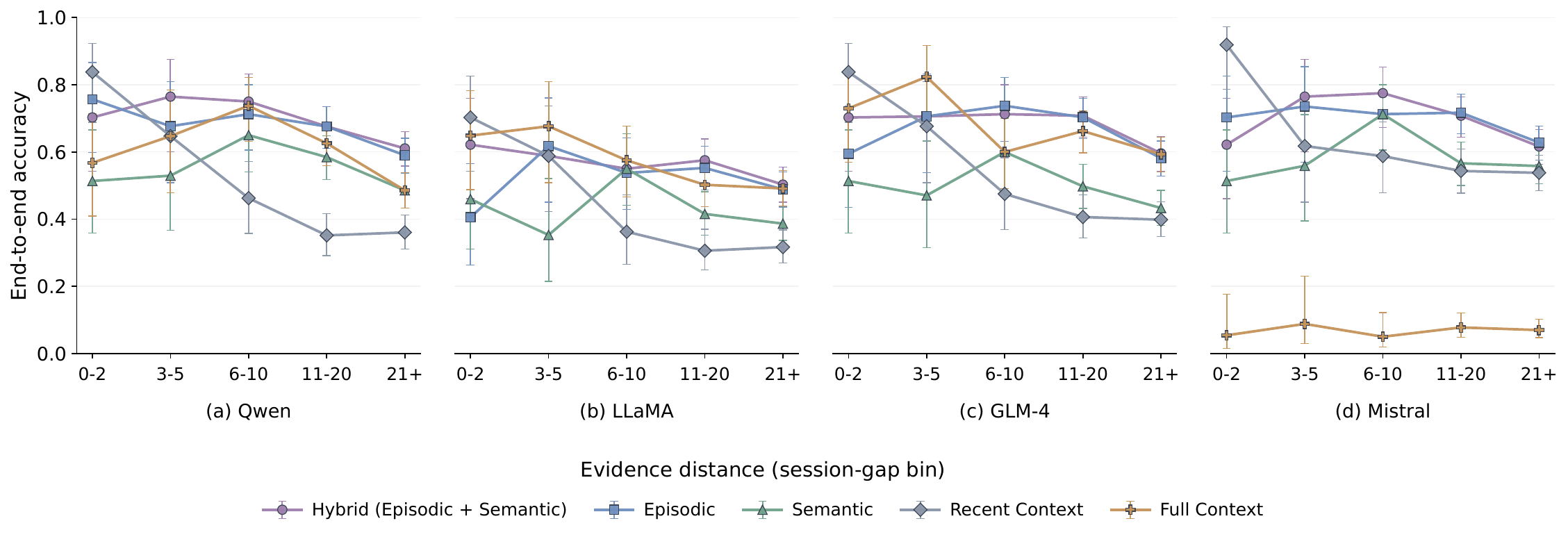}
    \caption {Accuracy across evidence-distance bins for each context strategy and answer model. Evidence distance is measured as the session gap between the query and its supporting evidence. Points indicate accuracy within each distance bin, with error bars showing 95\% Wilson confidence intervals.}
    \label{fig:rq2_distance}
\end{figure*}

\subsection{Effect of Query-Evidence Distance}

To examine the effect of evidence distance, we evaluate accuracy across five session-gap bins: 0--2, 3--5, 6--10, 11--20, and 21+ sessions. Figure~\ref{fig:rq2_distance} shows a pronounced decline for Recent Context from the 0--2 to the 21+ bin for Qwen (83.8\% to 36.1\%), LLaMA (70.3\% to 31.7\%), GLM-4 (83.8\% to 39.8\%), and Mistral (91.9\% to 53.8\%). At 21+ sessions, Hybrid achieves 61.1\%, 50.3\%, and 59.6\% for Qwen, LLaMA, and GLM-4, respectively, while Episodic achieves 62.8\% for Mistral.

We further fit a logistic regression of answer correctness on standardized session gap, context strategy, and their interaction, with cluster-robust standard errors by question. Under Episodic (the reference strategy), increasing distance is associated with lower correctness for Qwen ($\beta=-0.168$, $p=.030$), Mistral ($\beta=-0.199$, $p=.013$), and GLM-4 ($\beta=-0.200$, $p=.012$), but not LLaMA ($\beta=-0.084$, $p=.261$). Most strategy-by-distance interactions are not significant; the exception is Recent Context for LLaMA, which declines more steeply than Episodic ($\beta=-0.216$, $p=.028$).

As a sensitivity analysis, we repeated the distance-stratified evaluation using only successful generations. The decline in Recent Context persisted across all four models, and Episodic and Hybrid remained above Recent Context in the 21+ session bin, indicating that generation failures alone do not explain the observed distance effect.

\subsection{Abstention on Unanswerable Questions}
\label{sec:abstention}

Context strategy also affects how models respond when the available patient history does not support the requested conclusion. As shown in Table~\ref{tab:rq3_abstention}, averaged across the four evaluated models, accuracy on unanswerable questions is lowest for Full Context (14.8\%) and highest for Semantic retrieval (33.5\%). Hybrid achieves the highest accuracy on answerable questions (63.1\%) while maintaining relatively high accuracy on unanswerable questions (30.9\%). Episodic shows a similar pattern, with 61.9\% and 28.7\% accuracy on answerable and unanswerable questions, respectively.

Performance on unanswerable questions varies across models and context strategies. Semantic achieves the highest average unanswerable-question accuracy, but its answerable-question accuracy (50.2\%) is lower than that of Hybrid and Episodic. Moreover, the strategy with the highest unanswerable-question accuracy differs across models. Overall, these results show that context strategy influences both reasoning when supporting evidence is available and abstention when the requested conclusion is unsupported.

\begin{table}[!t]
\centering
\caption{Accuracy (\%) on answerable and adversarial questions. Adversarial questions contain unsupported premises or request conclusions that cannot be established from the available patient history. Mean denotes the average across the four answer models. Bold indicates the highest model-specific accuracy within each context strategy and question category.}
\label{tab:rq3_abstention}

\footnotesize
\renewcommand{\arraystretch}{1.05}

\begin{tabular*}{\linewidth}{@{\extracolsep{\fill}}llcc@{}}
\toprule
\textbf{Strategy} & \textbf{Model} &
\textbf{Answerable} &
\textbf{Adversarial} \\
\midrule

\multirow{5}{*}{Full}
& Qwen    & 56.9 & 16.8 \\
& Mistral & 7.0  & \textbf{17.1} \\
& LLaMA   & 52.1 & 12.9 \\
& GLM-4   & \textbf{63.3} & 12.2 \\
\cmidrule(l){2-4}
& Mean    & 44.8 & 14.8 \\
\midrule

\multirow{5}{*}{Recent}
& Qwen    & 40.8 & 8.7 \\
& Mistral & \textbf{56.9} & \textbf{59.1} \\
& LLaMA   & 35.2 & 21.3 \\
& GLM-4   & 44.5 & 33.6 \\
\cmidrule(l){2-4}
& Mean    & 44.3 & 30.7 \\
\midrule

\multirow{5}{*}{Episodic}
& Qwen    & 64.3 & 15.7 \\
& Mistral & \textbf{67.4} & \textbf{48.6} \\
& LLaMA   & 51.5 & 27.3 \\
& GLM-4   & 64.3 & 23.1 \\
\cmidrule(l){2-4}
& Mean    & 61.9 & 28.7 \\
\midrule

\multirow{5}{*}{Semantic}
& Qwen    & 53.8 & 18.5 \\
& Mistral & \textbf{57.6} & \textbf{62.2} \\
& LLaMA   & 41.6 & 26.9 \\
& GLM-4   & 47.8 & 26.2 \\
\cmidrule(l){2-4}
& Mean    & 50.2 & 33.5 \\
\midrule

\multirow{5}{*}{Hybrid}
& Qwen    & 65.8 & 19.6 \\
& Mistral & \textbf{66.9} & \textbf{47.6} \\
& LLaMA   & 54.1 & 25.9 \\
& GLM-4   & 65.4 & 30.8 \\
\cmidrule(l){2-4}
& Mean    & 63.1 & 30.9 \\

\bottomrule
\end{tabular*}
\end{table}

\subsection{Cross-Benchmark Evaluation on LongMemEval}
\label{sec:longmemeval}

To assess whether the effects of context strategy extend beyond longitudinal clinical histories, we additionally evaluate the five strategies on LongMemEval \cite{wu2025longmemeval}. LongMemEval is a benchmark for evaluating long-term conversational memory in LLMs, requiring models to retrieve and reason over information distributed across extended, multi-session interaction histories. This provides a complementary general-domain setting to MedLoCoMo while preserving the core challenge of identifying relevant evidence from long histories.

\begin{figure}{h}
    \centering
    \includegraphics[width=\linewidth]{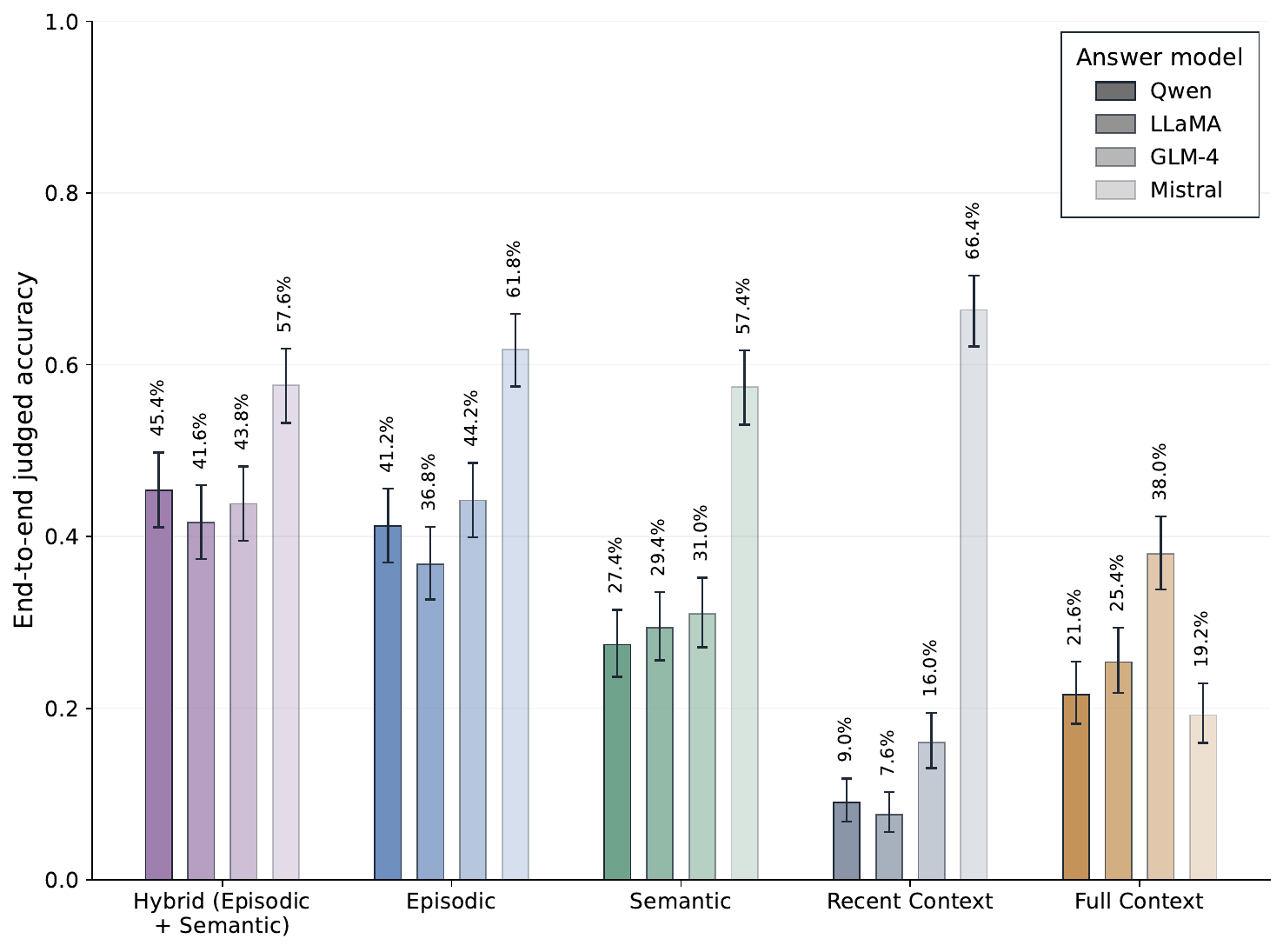}
    \caption{End-to-end judged accuracy on LongMemEval across context strategies and answer models. Error bars indicate 95\% Wilson confidence intervals.}
    \label{fig:longmemeval_results}
\end{figure}

Figure~\ref{fig:longmemeval_results} shows that context strategy continues to have a substantial effect on end-to-end accuracy, although the magnitude varies across models. Hybrid achieves 45.4\%, 41.6\%, 43.8\%, and 57.6\% for Qwen, LLaMA, GLM-4, and Mistral, respectively, while Episodic achieves 41.2\%, 36.8\%, 44.2\%, and 61.8\%. Semantic retrieval is particularly effective for Mistral (57.4\%) but lower for the other models (27.4-31.0\%). Recent Context performs poorly for Qwen (9.0\%), LLaMA (7.6\%), and GLM-4 (16.0\%), whereas Mistral reaches 66.4\%, illustrating substantial model-dependent variation. Full Context similarly varies across models, ranging from 19.2\% for Mistral to 38.0\% for GLM-4.

Overall, the cross-benchmark results support the broader finding that context selection materially affects long-horizon reasoning, while also showing that no single strategy behaves uniformly across models and domains.

\subsection{Discussion}

Our results show that providing more longitudinal history does not necessarily improve clinical reasoning. Episodic and Hybrid generally outperform Recent and Full Context, despite Full Context providing access to more of the patient history. This suggests that selectively retrieving relevant evidence can be more effective than simply increasing the amount of context available to the answer model.

This distinction becomes more apparent as evidence becomes more distant. Recent Context performs better when relevant evidence is nearby but degrades as the session gap increases, whereas Episodic and Hybrid retain higher accuracy at long distances, although their distance slopes do not differ significantly from other strategies. However, performance varies across reasoning tasks and answer models, indicating that the effectiveness of a context strategy depends on both the reasoning requirement and the underlying model.

Context strategy also influences how models handle questions with unsupported premises. Semantic achieves the highest average accuracy on adversarial questions, while Hybrid combines the highest average accuracy on answerable questions with relatively strong adversarial-question performance. The strategy with the highest adversarial-question accuracy nevertheless differs across models. Together, these findings show that context strategy affects multiple dimensions of long-horizon reasoning: answer correctness, performance on distant evidence, and abstention when the available history does not support the requested conclusion.

\section{Conclusion}
\label{sec:conclusion}
In this study, we investigated how context strategy affects long-horizon clinical reasoning by comparing Full, Recent, Episodic, Semantic, and Hybrid context across multiple open-weight language models. Episodic and Hybrid generally achieve stronger answer accuracy, including on distant evidence, while performance on adversarial questions reveals that successful reasoning when evidence is available does not necessarily translate to successful abstention when it is not. Overall, the results demonstrate that reliable longitudinal reasoning depends not only on how much historical information is available, but also on how relevant evidence is selected and presented to the model. This study has several limitations. We evaluate a fixed set of context strategies and retrieval settings (a single embedding model and a 7,000-token evidence budget) with four small open-weight models (3-9B parameters) served locally, so results may differ for larger models or other configurations. Semantic facts were extracted with Qwen3-4B, which shares a model family with one answer model. MedLoCoMo histories are synthetic dialogues derived from MIMIC-IV records and may not reflect the noise and heterogeneity of real clinical documentation. Finally, correctness was assessed by LLM judges whose agreement was moderate and was not validated against clinician annotations. Future work will investigate adaptive approaches that select or combine context strategies based on the query and reasoning requirements, and evaluate their reliability in more realistic clinical settings.

\section*{Acknowledgment}
The research reported in this publication was supported by the Predictive Analytics and Technology Integration (PATENT) Laboratory in the Department of Computer Science at The University of Alabama.

% \bibliographystyle{iclr2027_conference}

% \newpage
% \balance
\bibliographystyle{IEEEtran}
\bibliography{references}
\newpage
\end{document}